\def\year{2017}\relax
\documentclass[a4paper]{article}
\usepackage{aaai17}
\usepackage{times}
\usepackage{helvet}
\usepackage{courier}
\usepackage{ensuretext}
\usepackage{mathnotation}
\usepackage{amsfonts}
\usepackage{algorithm}
\usepackage{color}	
\definecolor{light-gray1}{gray}{0.95}
\usepackage[table]{xcolor}
\usepackage{amsthm}
\usepackage{url}
\usepackage{graphicx}
\usepackage{multirow}
\usepackage{balance}
\usepackage{flushend}

\newcommand{\mo}{\mathcal{K}}
\newcommand{\mb}{\mathcal{B}}
\newcommand{\md}{\mathcal{D}}
\newcommand{\mc}{\mathcal{C}}
\newcommand{\Tp}{\mathit{P}}
\newcommand{\Tn}{\mathit{N}}
\newcommand{\tp}{\mathit{p}}
\newcommand{\tn}{\mathit{n}}
\newcommand{\RQ}{{\mathit{R}}}
\newcommand{\ot}{\mathcal{K}^{*}}
\newcommand{\minD}{{\bf{mD}}}
\newcommand{\dx}[1]{{\bf D}_{#1}^+}
\newcommand{\dnx}[1]{{\bf D}_{#1}^{-}}
\newcommand{\dz}[1]{{\bf D}_{#1}^0}
\newcommand{\mQ}{{\bf{Q}}}
\newcommand{\mD}{{\bf{D}}}
\newcommand{\Pt}{\mathfrak{P}}
\newcommand{\Disc}{\mathsf{Disc}}

\newtheorem{definition}{Definition}{}

\nocopyright

\makeatletter
\def\@copyrightspace{\relax}
\makeatother
 
\begin{document}
%
\title{Scalable Computation of Optimized Queries for Sequential Diagnosis}

\author{Patrick Rodler \and Wolfgang Schmid \and Kostyantyn Shchekotykhin \\
	\multicolumn{1}{p{.7\textwidth}}{\centering\emph{Alpen-Adria Universit\"at, Klagenfurt, 9020 Austria}} \\ 
firstname.lastname@aau.at}

\maketitle
\begin{abstract}
In many model-based diagnosis applications it is impossible to provide such a set of observations and/or measurements that allow to identify the real cause of a fault. Therefore, diagnosis systems often return many possible candidates, leaving the burden of selecting the correct diagnosis to a user. Sequential diagnosis techniques solve this problem by automatically generating a sequence of queries to some oracle. The answers to these queries provide additional information necessary to gradually restrict the search space by removing diagnosis candidates inconsistent with the answers.

During query computation, existing sequential diagnosis methods often require the generation of many unnecessary query candidates and strongly rely on expensive logical reasoners. We tackle this issue by devising efficient heuristic query search methods. The proposed methods enable for the first time a completely reasoner-free query generation while at the same time guaranteeing optimality conditions, e.g. minimal cardinality or best understandability, of the returned query that existing methods cannot realize. Hence, the performance of this approach is independent of the (complexity of the) diagnosed system.
Experiments conducted using real-world problems show that the new approach is highly scalable and outperforms existing methods by orders of magnitude.
\end{abstract}

\section{Introduction}
Model-based diagnosis (MBD) is a principled approach to finding explanations, called diagnoses, for unexpected behavior of observed systems. MBD approaches are applied for, e.g.\, the diagnosis of hardware \cite{Reiter87,dekleer1987,dressler1996consistency,Wotawa:2001:DHD:590896.590976} and software \cite{DBLP:conf/aadebug/MateisSWW00,DBLP:conf/ijcai/StumptnerW99,DBLP:conf/aaai/ElmishaliSK16}, knowledge bases \cite{friedrich2005gdm,Kalyanpur.Just.ISWC07}, feature models \cite{DBLP:journals/jss/WhiteBSTDC10}, discrete event systems \cite{pencole2005formal,DBLP:conf/ijcai/SuZG16}, user interfaces \cite{DBLP:journals/apin/FelfernigFISTJ09}, etc.

In many cases, however, the observations of a faulty system provide insufficient information for successful fault localization. Therefore, MBD methods might return many possible  diagnoses and a user must choose the correct diagnosis manually. Given the complexity of modern systems, selection of the correct fault explanation is a hard task.

Sequential diagnosis methods, e.g.\ \cite{dekleer1987,Siddiqi2011,pietersma2005model}, allow a user to find the correct diagnosis by answering a sequence of queries. 
Every query is constructed in a way that, given any possible answer of the user, at least one diagnosis is inconsistent with it. Consequently, in every iteration of a sequential method at least one diagnosis is pruned. This approach guarantees identification of the correct diagnosis after a finite number of iterations.

Existing sequential methods rely on strategies for query computation which often lead to the generation of many unnecessary query candidates and thus to a high number of expensive calls to reasoning services. Such an overhead can have a severe impact on the response time of the interactive debugger, i.e.\ the time between two consecutive queries. Problems of these approaches are (1)~the necessity to actually \emph{compute} a query candidate in order to asses its goodness, (2)~the very expensive verification whether a candidate is a query, (3)~their inability to recognize candidates that are no queries before verification, (4)~the possibility of the generation of query duplicates and (5)~the strong dependence on the output of used reasoning services regarding the completeness and quality of the set of queries generated. 

To tackle these issues, we devise an efficient heuristic query computation algorithm that solves all the mentioned problems of legacy approaches. Our main contributions are: 
\begin{itemize}
    \item We introduce a new, theoretically well-founded and sound method for query generation that works completely \emph{without the use of reasoning services}. 
    
    \item Our algorithm guarantees optimality of each query w.r.t.\ (a) the expected number of subsequent queries and (b) required properties, e.g.\ the number of elements in a query.
    
    \item Given a preference criterion, the method can reformulate a query to comprise elements considered simple by a user.  
    
    
    \item Evaluation results show the quasi constant execution time of our method for increasing problem size and the significant superiority to existing methods throughout all tests, e.g.\ for problems with $\geq 15$ diagnoses our method always needs less than 1\% of the time existing methods require.
\end{itemize}

To summarize, the proposed query generation method produces a query that is optimized along different dimensions whereas its runtime accounts for only a minor fraction of the reaction time, i.e.\ the time interval between two queries of a sequential diagnosis method.

\section{Preliminaries}
%
%
%
In this section we review the basics of sequential diagnosis methods. 
In particular, the approach described in this paper is a variant of a sequential diagnosis technique suggested in \cite{Shchekotykhin2012}. 
The latter considers the diagnosis problem wrt.\ interactive KB debugging\footnote{See \cite{Rodler2015phd} for a comprehensive treatment.} which, however, can be generalized to various other MBD approaches.

As to the notation, we will write $\mo \models X$ for sets of logical formulas $\mo$ and $X$ to denote that $\mo \models x$ for all $x \in X$. Moreover, $U_X$ and $I_X$ for some collection of sets $X$ refers to the union and the intersection of all sets in $X$, respectively. In addition, we assume that entailment ($\models$) is monotonic.

\subsubsection{Knowledge Base Debugging.} A \emph{(parsimonious) KB debugging problem (KDP)} involves finding a maximal solution KB for a diagnosis problem instance given as input.
In a \emph{diagnosis problem instance (DPI)} $\tuple{\mo,\mb,\Tp,\Tn}_\RQ$ over some logic $\mathcal{L}$, $\mo$ is a KB of potentially faulty formulas, $\mb$ a (background) KB of formulas assumed as correct. Further, $\RQ \supseteq \setof{\text{consistency}}$ is a set of KB quality requirements a solution KB must meet. $\Tp$ and $\Tn$ are sets of positive and negative test cases, respectively. Each test case is a set (conjunction) of formulas.
All elements of $\mo$, $\mb$ and of each test case are formulated over the logic $\mathcal{L}$.
A KB $\ot$ over $\mathcal{L}$ is a \emph{solution KB} for this DPI iff $\ot \cup \mb$ meets all requirements $r\in\RQ$, $\ot \cup \mb \models \tp$ for all $\tp \in \Tp$ and $\ot \cup \mb \not\models \tn$ for all $\tn\in\Tn$. It is a \emph{maximal solution KB} iff there is no solution KB $\mo'$ such that $\mo' \cap \mo \supset \ot\cap\mo$, i.e.\ the intersection with the given KB $\mo$ must be $\subseteq$-maximal. 
A KB $\mo$ is said to be \emph{faulty} w.r.t.\ $\tuple{\cdot,\mb,\Tp,\Tn}_\RQ$ iff $\mo \cup \mb \cup U_\Tp$ violates some $r\in\RQ$ or entails some $\tn\in\Tn$.
A KB $\mc$ is a \emph{conflict set} for $\tuple{\mo,\mb,\Tp,\Tn}_\RQ$ iff $\mc \subseteq \mo$ and $\mc$ is faulty w.r.t.\ $\tuple{\cdot,\mb,\Tp,\Tn}_\RQ$. A conflict set $\mc$ is minimal iff there is no conflict set $\mc'$ such that $\mc'\subset\mc$.
A set of formulas $\md \subseteq \mo$ is called a \emph{diagnosis} for $\tuple{\mo,\mb,\Tp,\Tn}_\RQ$ iff $(\mo\setminus\md)\cup U_\Tp$ is a solution KB. A diagnosis $\md$ is \emph{minimal} iff there is no diagnosis $\md'$ such that $\md' \subset \md$. The task of a KDP reduces to computing a minimal diagnosis for the given DPI \cite[Prop.~3.6]{Rodler2015phd}.
The set of all minimal diagnoses for a DPI $\mathsf{DPI}$ are henceforth denoted by $\minD(\mathsf{DPI})$. The relation between minimal conflict sets and minimal diagnoses is as follows \cite{Reiter87}:  A (minimal) diagnosis for a DPI $\mathsf{DPI}$ is a (minimal) hitting set of all
minimal conflict sets for $\mathsf{DPI}$.

\begin{table}[t]
	\scriptsize
	\centering
		\begin{tabular}{llll} 
			\hline
			\multirow{2}{*}{$\mo$}  & $\phi_1: A \rightarrow B \land L$ & $\phi_2: A \to F$ & $\phi_3: B \lor F \to H$  \\ 
															& $\phi_4: L \to H$ 								& $\phi_5:\lnot H \to G \land \lnot A$ & 	\\
			\hline
			$\Tn$  &  \multicolumn{1}{l|}{$\tn_1 : \setof{A \to H}$} &   $\RQ =\setof{\text{consistency}}$ & \multicolumn{1}{|l}{$\Tp, \mb = \emptyset$} \\
			\hline
			\end{tabular}
	\caption{Propositional Logic Example DPI $\mathsf{DPI_{ex}}$} 
	\label{tab:example_dpi_1}
\end{table}

\vspace{2pt}

\noindent\textbf{Example:} To illustrate these notions, consider the example DPI $\mathsf{DPI}_{\mathsf{ex}}$ given by Table~\ref{tab:example_dpi_1}. Denoting KB formulas $\phi_i$ simply by $i$, 
the set of all minimal conflict sets for $\mathsf{DPI}_{\mathsf{ex}}$ is $\setof{\setof{1,3},\setof{1,4},\setof{2,3},\setof{5}}$. This holds since each of these sets entails the negative test case $\tn_1 = \setof{A \to H}$. Further, the set of all minimal diagnoses $\minD(\mathsf{DPI_{ex}}) = \setof{\md_1,\md_2,\md_3} = \setof{\setof{1,2,5},\setof{1,3,5},\setof{3,4,5}}$ (minimal hitting sets of all minimal conflict sets). 
All maximal solution KBs are given by $\setof{3,4}$, $\setof{2,4}$ and $\setof{1,2}$. Note that e.g.\ both the empty KB $\emptyset$ and $\setof{X \to Y}$ are also solution KBs since both are consistent (meet $r_1 \in \RQ$) and do not entail $\tn_1 \in \Tn$, albeit not maximal ones.\qed

\subsubsection{Interactive Knowledge Base Debugging.} For one and the same DPI there can be a large number of different (maximal) solution KBs. Each of them has different semantics. Interactive KB debugging aims at restricting the solution space until a single solution (with exactly the desired semantics) is left. Given a DPI $\tuple{\mo,\mb,\Tp,\Tn}_\RQ$ as input, an \emph{interactive (dynamic) KB debugging problem (IKDP)} involves finding a maximal solution KB $\ot$ for a DPI $\tuple{\mo,\mb,\Tp',\Tn'}_\RQ$ where $\Tp' \supseteq \Tp$, $\Tn' \supseteq \Tn$ such that $\ot$ is the only maximal solution KB for this DPI. 
That is, solving the IKDP means specifying a set $\Tp' \setminus \Tp$ of new positive and a set $\Tn' \setminus \Tn$ of new negative test cases until a single minimal diagnosis is left or, in a relaxed version of IKDP, some minimal diagnosis has a probability higher than some threshold \cite{Shchekotykhin2012}. 
These new test cases are queries the debugger poses to an interacting user. A query $Q$ is a $t$-$f$-question if the set of formulas constituted by it must ($t$) or must not ($f$) be true in the intended domain (i.e.\ entailed by the correct solution KB). An answered query leads to a new DPI, i.e.\ a positive answer causes the addition of $Q$ to $\Tp$, a negative one to $\Tn$. 

To be a query, a set of formulas must satisfy two conditions, (1)~invalidate at least one minimal diagnosis (search space restriction) and (2)~preserve the validity of at least one minimal diagnosis (solution preservation). 
Formally: A set $Q \neq \emptyset$ of logical formulas over $\mathcal{L}$ is a \emph{query} for the DPI $\tuple{\mo,\mb,\Tp,\Tn}_\RQ$ iff $\minD({\tuple{\mo,\mb,\Tp,\Tn}_\RQ})\setminus\minD(\tuple{\mo,\mb,\Tp\cup\setof{Q},\Tn}_\RQ) \neq \emptyset$ and $\minD(\tuple{\mo,\mb,\Tp,\Tn}_\RQ)\setminus\minD(\tuple{\mo,\mb,\Tp,\Tn\cup\setof{Q}}_\RQ) \neq \emptyset$, i.e.\ both answers eliminate at least one minimal diagnosis. That is, at least two minimal diagnoses are required to test whether a set of formulas $Q$ is a query. Since the calculation of the entire set $\minD({\tuple{\mo,\mb,\Tp,\Tn}_\RQ})$ is generally not tractable within reasonable time due to the complexity of diagnosis computation \cite[Sec.~9.4]{Rodler2015phd}, usually a fixed-cardinality subset 
$\mD$ of it,
the \emph{leading diagnoses}, are exploited. In most cases, the latter are the minimal diagnoses of minimal cardinality or maximal probability.
Given a DPI $\tuple{\mo,\mb,\Tp,\Tn}_\RQ$, we denote by $\mQ_{\mD,\tuple{\mo,\mb,\Tp,\Tn}_\RQ}$ all queries $Q$ for $\tuple{\mo,\mb,\Tp,\Tn}_\RQ$ which satisfy $\mD\setminus\minD(\tuple{\mo,\mb,\Tp\cup\setof{Q},\Tn}_\RQ) \neq \emptyset$ and $\mD\setminus\minD(\tuple{\mo,\mb,\Tp,\Tn\cup\setof{Q}}_\RQ) \neq \emptyset$, i.e.\ both answers eliminate at least one \emph{leading} diagnosis. For brevity and when the DPI is clear, we will often write only $\mQ_{\mD}$.

Let $\mo_i^* := (\mo\setminus\md_i)\cup U_{\Tp} \cup \mb$ for each $\md_i \in \minD(\tuple{\mo,\mb,\Tp,\Tn}_\RQ)$, i.e.\ $\mo_i^*$ is the solution KB resulting from the deletion of $\md_i$ along with the background KB. Each query candidate (i.e.\ set of formulas $Q\neq \emptyset$) partitions the set $\mD$ into three parts. These are $\dx{}(Q) = \setof{\md_i \in \mD\,|\,\mo_i^* \models Q}$, i.e.\ the leading diagnoses consistent only with $Q$'s positive answer, $\dnx{}(Q) = \setof{\md_i \in \mD\,|\,\exists x \in \RQ\cup\Tn: \mo_i^*\cup Q \text{ violates }x}$, i.e.\ the leading diagnoses consistent only with $Q$'s negative answer, and $\dz{}(Q) = \mD \setminus (\dx{}(Q) \cup \dnx{}(Q))$, i.e.\ the leading diagnoses consistent with both answers. The tuple $\Pt(Q) := \tuple{\dx{}(Q),\dnx{}(Q),\dz{}(Q)}$ is called the \emph{q-partition (QP)} of $Q$ for $\mD$ iff $Q$ is a query, i.e.\ iff $Q \in \mQ_\mD$. Thus, not each partition of $\mD$ into three parts is necessarily a QP. Given a QP $\Pt$, we sometimes call its three entries in turn $\dx{}(\Pt)$, $\dnx{}(\Pt)$ and $\dz{}(\Pt)$. $Q$ is called \emph{query with (or: for) the QP} $\Pt$ iff $\Pt$ is the QP of $Q$. In general \cite[Prop.~56]{DBLP:journals/corr/Rodler16a}, there can be exponentially many (non-equivalent) queries for a single QP. 
Each query $Q$ is a subset of the common entailments of all KBs in the set $\setof{\mo^*_i\,|\,\md_i \in \dx{}(Q)}$.

The sets $\dx{}$ and $\dnx{}$ are a helpful instrument in deciding whether a set of formulas is a query or not. Namely, a non-empty set of formulas $X$ is in $\mQ_\mD$ iff $\dx{}(X) \neq \emptyset$ and $\dnx{}(X) \neq \emptyset$. That is, for all queries $Q \in \mQ_\mD$, the first two entries of $\Pt(Q)$ must be non-empty. 
Moreover, $\Pt(Q)$ facilitates an estimation of the impact $Q$'s answers have in terms of the invalidation of minimal diagnoses. And, given (diagnosis) fault probabilities, it enables to gauge the probability of getting a positive or negative answer to a query. Active learning query selection measures $m: Q \mapsto m(Q) \in \mathbb{R}$ \cite{settles2012,dekleer1987,Shchekotykhin2012,Rodler2013} use exactly these query properties characterized by the QP to assess how favorable a query is. 

\vspace{2pt}

\noindent\textbf{Example (cont'd):} Let the set of leading diagnoses $\mD = \minD(\mathsf{DPI_{ex}}) = \setof{\md_1,\md_2,\md_3}$. Then, $Q = \setof{F \to H}$ is a query in $\mQ_\mD$. To verify this, let us consider its QP $\Pt(Q) = 
\tuple{\setof{\md_1},\setof{\md_2,\md_3},\emptyset}$. Since both $\dx{}(Q)$ and $\dnx{}(Q)$ are non-empty, $Q$ is in $\mQ_\mD$. $\md_1 \in \dx{}(Q)$ holds as $\mo^*_1 = (\mo \setminus \md_1) \cup \mb \cup U_\Tp = (\setof{1,\dots,5}\setminus\setof{1,2,5}) \cup \emptyset \cup \emptyset = \setof{3,4} = \setof{B\lor F \to H, L\to H}$ entails $Q$. On the other hand, e.g.\ $\md_2 \in \dnx{}(Q)$ due to the fact that 
$\mo^*_2 \cup Q = \setof{A\to F, L \to H, F \to H} \models \setof{A \to H} = \tn_1 \in \Tn$. 
Hence, $Q$'s positive answer implies that leading diagnoses in $\dnx{}(Q) = \setof{\md_2,\md_3}$ are invalidated, i.e.\ are not in $\minD(\tuple{\mo,\mb,\Tp\cup\setof{Q},\Tn}_\RQ)$. Likewise, a negative answer causes $\setof{\md_1} \cap \minD(\tuple{\mo,\mb,\Tp,\Tn\cup\setof{Q}}_\RQ) = \emptyset$.\qed

\section{Heuristic-Based Query Optimizer (H-QUO)}
In this section, we propose a new implementation of the query selection in interactive KB debugging.\footnote{The new query computation mechanism H-QUO has already been realized in terms of a Prot\'eg\'e plug-in for sequential/interactive ontology debugging, see \url{http://isbi.aau.at/ontodebug/}. Prot\'eg\'e is currently the most widely used open-source ontology engineering software in the world, see \url{http://protege.stanford.edu/}.} Before, we briefly discuss issues that come with existing approaches:

\subsubsection{Related Work.} 
To find a query $Q$ with (nearly) optimal value $m(Q)$ w.r.t.\ a measure $m$ given leading diagnoses $\mD$, existing interactive KB debuggers \cite{Shchekotykhin2012,Shchekotykhin2014,Rodler2013,Rodler2015phd} proceed as follows: They (1)~use a reasoner $rsnr$ to compute for different seeds $\dx{} \subset \mD$ a set $X$ of common entailments of all KBs in $\setof{\mo^*_i\,|\,\md_i \in \dx{}}$ and (2)~if $X \neq \emptyset$, then they use $rsnr$ to assign all diagnoses in $\mD \setminus \dx{}$ to their respective set among $\dx{}(X)$, $\dnx{}(X)$ and $\dz{}(X)$. (3)~If both $\dx{}(X)$ and $\dnx{}(X)$ are non-empty (i.e.\ $X$ is a query with QP $\Pt(X)$) and $m(X)$ is sufficiently good, a $\subseteq$-minimal subset $X'$ of $X$ is computed using $rsnr$ such that $\Pt(X') = \Pt(X)$.

In general, steps 1 and 2 are executed $O(2^{|\mD|})$ times 
yielding a worst case exponential number of expensive reasoner calls. \cite{Shchekotykhin2012} couples this process with a heuristic to direct the search faster towards optimal queries, but partitions suggested by the heuristic might in fact be no QPs which hinders efficient search space pruning. Also, the strong reasoner dependence persists. 

As pointed out by \cite{Rodler2015phd}, all these techniques suffer from the problem that the (number and types of) entailments output by $rsnr$, i.e.\ the shape of $X$, have a significant influence on the quality and the number of different calculated queries. \cite{Rodler2015phd} shows that objectively worse queries
might be found at the cost of neglecting better ones. Further drawbacks are that duplicate QPs might occur in the search (as different seeds can lead to same QP), effective employment of heuristics and pruning is not possible (see above) and that no guarantees whatsoever can be given w.r.t.\ properties
of the minimized query $X'$. The method H-QUO proposed in this work solves all these issues.

\subsubsection{The New Approach.} We propose a four-phase optimization of a query
(for an in-depth discussion, all algorithms and proofs see \cite{DBLP:journals/corr/Rodler16a}). In the first phase (P1), an optimal QP $\Pt$ w.r.t.\ $m$ is computed (\textsc{findQPartition}). Second (P2), a best query $Q$ for $\Pt$ w.r.t.\ a secondary criterion, e.g.\ minimum cardinality, is computed (\textsc{selectQueryForQPartition}). Third (P3), $Q$ is enriched by formulas of simple structure, e.g.\ simple implications of the form $A \to B$ for Propositional Logic, resulting in $Q' \supseteq Q$ (\textsc{enrichQuery}). Fourth (P4), $Q'$ is minimized and optimized in a way that, roughly, most of the simple formulas are maintained and most of the more complex ones deleted, yielding the final query $Q^*$ (\textsc{optimizeQuery}). Note P3 and P4 are optional and require altogether a polynomial number of reasoner calls. P1 and P2 do not require any reasoner. Next, we give the main ideas of the four functions.

\subsubsection{P1 -- \textsc{findQPartition}.} 
To overcome the mentioned problems of existing systems, we use the notion of a canonical query, a well-defined subset of $\mo$. 
Any query $Q \subseteq \mo$ in $\mQ_\mD$ \emph{must} include some formulas in $U_\mD$, \emph{need not} include any formulas in $\mo \setminus U_\mD$, and \emph{must not} include any formulas in $I_\mD$ \cite[Cor.~19]{DBLP:journals/corr/Rodler16a}. Moreover, the deletion of any formulas in $\mo \setminus U_\mD$ from $Q$ does not alter the QP $\Pt(Q)$. Hence, we call $\Disc_\mD := U_\mD \setminus I_\mD$ the discrimination formulas (i.e.\ those that are essential in discriminating between leading diagnoses $\mD$). 
\begin{definition}
Let $\emptyset\subset\dx{}\subset\mD$. Then we call $Q_{\mathsf{can}}(\dx{}) := (\mo \setminus U_{\dx{}}) \cap \Disc_\mD$ \emph{the canonical query (CQ)} w.r.t.\ $\dx{}$ if $Q_{\mathsf{can}}(\dx{}) \neq \emptyset$. Otherwise, $Q_{\mathsf{can}}(\dx{})$ is undefined.
\end{definition}
In that, $\mo \setminus U_{\dx{}}$ are exactly the common (explicit) entailments of $\setof{\mo \setminus \md_i\,|\,\md_i \in \dx{}}$, i.e.\ the CQ is a minimization of this set under preservation of the QP. 
%
\begin{definition}
A QP $\Pt'$ for which a CQ $Q$ exists with exactly this QP, i.e.\ $\Pt(Q) = \Pt'$, is called a \emph{canonical QP (CQP)}.
\end{definition} 
A restriction -- only for now, during P1 -- to searching only for CQs, has some nice implications: (1)~CQs can be generated by cheap set operations (no reasoner calls), (2)~each CQ is a query in $\mQ_\mD$ for sure, no verification of its QP required, thence no unnecessary 
query candidates tested,
(3)~automatic focus on favorable queries (those with empty $\dz{}$), 
(4)~no duplicate QPs as there is a one-to-one relationship between CQs and CQPs. 

%

\vspace{2pt}

\noindent\textbf{Example (cont'd):} 
Let $\mD$ as before. Then $\Disc_\mD = U_\mD \setminus I_\mD = \setof{1,2,3,4,5}\setminus\setof{5} = \setof{1,2,3,4}$. Let us consider the seed $\dx{} = \setof{\md_1}$. Then the CQ $Q_1 := Q_{\mathsf{can}}(\dx{}) = (\setof{1,\dots,5} \setminus \setof{1,2,5}) \cap \setof{1,\dots,4} = \setof{3,4}$. The associated CQP is $\Pt_1 = \tuple{\setof{\md_1},\setof{\md_2,\md_3},\emptyset}$. Note $\md \in \dx{}(\Pt_1)$ ($\md \in \dnx{}(\Pt_1)$) for some $\md \in \mD$ iff $\mo \setminus \md \supseteq (\not\supseteq) Q_1$. E.g.\ $\md_3\in\dnx{1}$ since $(\mo \setminus \md_3) = \setof{1,2} \not\supseteq \setof{3,4} = Q_1$. That is, reasoning is traded for set operations and comparisons.

The seed $\dx{} = \setof{\md_1,\md_3}$ yields $Q_2 := Q_{\mathsf{can}}(\dx{})=(\setof{1,\dots,5} \setminus \setof{1,\dots,5}) \cap \setof{1,\dots,4} = \emptyset$, i.e.\ there is no CQ w.r.t.\ $\dx{}$ and the partition $\tuple{\setof{\md_1,\md_3},\setof{\md_2},\emptyset}$ with the seed as first entry is no CQP (and also no QP).
\qed 

\vspace{2pt}

Since the evaluation of $m(Q)$ for a query $Q$ requires the QP $\Pt(Q)$, we propose a (heuristic) depth-first, local best-first (i.e.\ chooses only among best direct successors at each step) backtracking search procedure that abstracts from queries and focuses only on QPs, in particular only on CQPs (and thence on CQs). The optimal query for the found QP is determined separately in the next phase P2. Note this separation is not possible without the notion of a CQ.

A search problem \cite{russellnorvig2010} 
is defined by
the \emph{initial state}, a \emph{successor function} enumerating all direct neighbor states of a state, the \emph{step costs} from a state to a successor state, some \emph{heuristics} to estimate the remaining effort towards a goal state, and the \emph{goal test} to determine if a given state is a goal state or not. We define the initial state as $\langle\emptyset,\mD,\emptyset\rangle$ (note this is not a QP). The idea is to transfer diagnoses step-by-step from $\dnx{}$ to $\dx{}$ to construct all CQPs systematically. The step costs are irrelevant, only the found QP as such counts, not the way to get there. Heuristics can be formulated based on qualitative requirements on optimal queries which can be derived from the real-valued function $m$ and used for search space pruning (as shown for numerous measures in \cite{DBLP:journals/corr/Rodler16a}). A QP is considered a goal if it optimizes $m$ up to some threshold $t$ (cf.\ \cite{Shchekotykhin2012}). In order to characterize a suitable successor function, we define a direct neighbor of a QP by means of:
	\begin{definition}\label{def:minimal_transformation}
	Let $\Pt_i := \langle \dx{i},\dnx{i},\emptyset\rangle$,
	$\Pt_j := \langle \dx{j},\dnx{j},\emptyset\rangle$ be partitions of $\mD$. 
	Then, $\Pt_i \mapsto \Pt_j$ is a \emph{minimal $\dx{}$-transformation} from $\Pt_i$ to $\Pt_j$ iff $\Pt_j$ is a CQP, $\dx{i} \subset \dx{j}$ and there is no CQP $\langle \dx{k},\dnx{k},\emptyset\rangle$ with $\dx{i} \subset \dx{k} \subset \dx{j}$.
	\end{definition}
From now on, we call a CQP $\Pt'$ a successor of a partition $\Pt$ iff $\Pt'$ results from $\Pt$ by a minimal $\dx{}$-transformation. So, all successors of the initial state have the form $\tuple{\setof{\md_i},\mD\setminus\setof{\md_i},\emptyset}$ \cite[Cor.~20]{DBLP:journals/corr/Rodler16a}. To specify successors of an intermediate CQP $\Pt_k$, we draw on traits:
\begin{definition}
Let $\Pt_k = \langle \dx{k}, \dnx{k}, \emptyset\rangle$ be a CQP and $\md_i \in \dnx{k}$. Then the trait $\md_i^{(k)}$ of $\md_i$ is defined as $\md_i \setminus U_{\dx{k}}$. 
\end{definition}
The relation $\sim_k$ associating two diagnoses in $\dnx{k}$ iff their trait is equal is an equivalence relation. Now, all successors of $\Pt_k$ have the form $\tuple{\dx{k} \cup E,\dnx{k}\setminus E,\emptyset}$ where $E$ is an equivalence class with a $\subseteq$-minimal trait among all equivalence classes $\setof{E_1,\dots,E_s}$ w.r.t.\ $\sim_k$ and $s \geq 2$. And, there are no other successors of $\Pt_k$. Usage of this successor function makes the search for CQPs complete. 
For the number $n_{\mathsf{CQP}}$ of CQPs it holds $n_{\mathsf{CQP}} = |\setof{U_{\dx{}}\,|\,\emptyset\subset\dx{}\subset\mD, U_{\dx{}} \neq U_\mD}| \geq |\mD|$ \cite[Cor.~22]{DBLP:journals/corr/Rodler16a}.
It is an open question whether QPs of the form $\tuple{\dx{},\dnx{},\emptyset}$ exist which are no CQPs, but \cite[Sec.~3.4.2]{DBLP:journals/corr/Rodler16a} gives theoretical and empirical evidence that, if there are such, then they should be quite rare. Moreover, our evaluation manifests that there are large numbers of CQPs in relation to $|\mD|$ which grants the detection of optimal QPs w.r.t.\ all measures $m$ discussed in the KB debugging literature \cite{Shchekotykhin2012,Rodler2013}.

\vspace{2pt}

\noindent\textbf{Example (cont'd):} Reconsider the CQP $\Pt_1$.
The traits are $\md_2^{(1)} = \setof{1,3,5} \setminus \setof{1,2,5} = \setof{3}$ and $\md_3^{(1)} = \setof{3,4}$, representing two equivalence classes w.r.t.\ $\sim_1$. There is only one class with $\subseteq$-minimal trait, i.e.\ $\setof{\md_2}$. Hence, there is a single successor CQP $\Pt_2 = \tuple{\setof{\md_1,\md_2},\setof{\md_3},\emptyset}$ of $\Pt_1$. Recall, we argued that $\tuple{\setof{\md_1,\md_3},\setof{\md_2},\emptyset}$ is indeed no CQP. 
\qed 
	

\subsubsection{P2 -- \textsc{selectQueryForQPartition}.} The QP $\Pt_k$ returned by P1 is already optimal w.r.t.\ the (e.g.\ information theoretic) measure $m$. Such measures aim at 
the minimization of subsequent queries until the correct diagnosis is found. Also, we have a CQ $Q_k$ for $\Pt_k$. However, usually a minimal requirement is the $\subseteq$-minimality of a query 
to reduce the effort for the user. To this end, let $\mathsf{Tr}(\Pt_k)$ denote the set of all $\subseteq$-minimal traits w.r.t.\ $\sim_k$. Then, $Q \subseteq \Disc_\mD$ is a $\subseteq$-minimal query with QP $\Pt_k$ iff $Q = H$ for some minimal hitting set of $\mathsf{Tr}(\Pt_k)$ \cite[Cor.~24]{DBLP:journals/corr/Rodler16a}.

Hence, all $\subseteq$-minimal reductions of CQ under preservation of the (already fixed and optimal) QP $\Pt_k$ can be computed e.g.\ using the classical $\textsc{HS-Tree}$ \cite{Reiter87}. However, there is a crucial difference to standard application scenarios of $\textsc{HS-Tree}$, namely the fact that all sets to label the tree nodes (i.e.\ the $\subseteq$-minimal traits) are readily available (without further computations). Consequently, the construction of the tree runs swiftly, as our evaluation will confirm (note also, in principle we only require a single minimal hitting set). Moreover, $\textsc{HS-Tree}$ can be used as uniform-cost search (cf.\ e.g.\ \cite[Chap.~4]{Rodler2015phd}), incorporating various quality criteria $crit$ such as minimum cardinality or best comprehensibility (by means of given fault information \cite[Sec.~3.5]{DBLP:journals/corr/Rodler16a}) of the output query. No existing approach can realize such optimizations. So far, P1 and P2 have computed -- without any call to a reasoner -- an optimal query $Q$ w.r.t.\ $m$ (\# of queries) and $crit$ (effort per query) which can be directly shown to the user. Optionally, $Q$ can be further enhanced in phases P3 and P4. 

\vspace{2pt}

\noindent\textbf{Example (cont'd):} Recall the CQP $\Pt_1$. Then, $\mathsf{Tr}(\Pt_1) = \setof{\setof{3}}$, i.e.\ there is a single minimum cardinality query $\setof{3}$ for $\Pt_1$, a proper subset of the CQ $\setof{3,4}$ for $\Pt_1$. 
Considering the CQP $\Pt_3 := \tuple{\setof{\md_2},\setof{\md_1,\md_3},\emptyset}$, we have $\mathsf{Tr}(\Pt_3) = \setof{\setof{2},\setof{4}}$ and thus a single query $\setof{2,4}$ of minimal size which is equal to the CQ for $\Pt_3$.  \qed

\subsubsection{P3 -- \textsc{enrichQuery}.} 
Sometimes a user might find it hard to classify formulas occurring explicitly in the KB as true or false statements in the intended domain, 
e.g.\ users are often convinced of the correctness of formulas they specified themselves.
In such a case it might be easier, less error-prone and more convenient for users to be presented with a query including \emph{simple} formulas not in the KB. In case of e.g.\ OWL or Description Logics, such formulas might be subsumption or class assertion axioms \cite{Baader2007}.

Therefore, \textsc{enrichQuery} realizes the expansion of the given query $Q \subseteq \mo$ returned by P2 by (finitely many) additional formulas 
$Q_{impl}$. 
At this, we postulate that (1)~$Q_{impl} \cap \mo \cup \mb \cup U_{\Tp} = \emptyset$ (only implicit entailments), (2)~ $S \models Q_{impl}$ where $S$ is some solution KB for the given DPI and $Q \subseteq S \subseteq \mo \cup \mb \cup U_{\Tp}$ (entailed by a consistent, non-faulty set of formulas) and (3)~no $\phi_i \in Q_{impl}$ is an entailment of $S \setminus Q$ (logical dependence on $Q$, no irrelevant formulas). Further on, we require that (4)~the query expansion does not alter the (already fixed and optimal) q-partition.
%

We define the expansion of $Q$ resulting in $Q' \gets Q \cup Q_{impl}$ as $Q_{impl} := [F^{(+Q)} \setminus F^{(-Q)}] \setminus Q$ where 
\small
\begin{align*} 
F^{(+Q)} &:= Ent_T[(\mo\setminus U_{\mD}) \cup \mb \cup U_{\Tp} \cup Q] \\
F^{(-Q)} &:= Ent_T[(\mo\setminus U_{\mD}) \cup \mb \cup U_{\Tp}] 
\end{align*}
\normalsize
and $Ent_T()$ is the function computing entailments of predefined formula types $T$ implemented by a reasoner. 
$Q_{impl}$ specified this way satisfies all postulated requirements (1) -- (3) \cite[Prop.~57]{DBLP:journals/corr/Rodler16a} and $Q'$ meets requirement (4) \cite[Prop.~58]{DBLP:journals/corr/Rodler16a}. Note that only two reasoner calls are required by the query enrichment. 

\subsubsection{P4 -- \textsc{optimizeQuery}.}
The enriched query $Q'$ returned by P3 is optimized in this phase. The plausible assumption now is that none of the simple formulas in $Q_{impl}$ is harder to understand than any formula in $Q$. And, the higher the fault probability of a formula is, the harder it is to understand for the user. The objective of \textsc{optimizeQuery} is 
(1)~the q-partition-preserving minimization of $Q'$ to obtain a $\subseteq$-minimal subset $Q^{*}$ of $Q'$ (low effort for user), 
(2)~$Q^{*} \cap Q = \emptyset$ (maintain only simple formulas), if such a $Q^{*}$ exists, and otherwise
(3)~if formula fault probabilities $p(\phi)$ for $\phi\in\mo$ are given, then $Q^{*}$ is required to be the one query among all $\subseteq$-minimal subsets of $Q'$ that minimizes the maximum probability $p(\phi)$ over all $\phi\in Q$ occurring in it (make hardest formula as simple as possible). 

Requirement (1) can be easily achieved by applying to $Q'$ the q-partition-preserving divide-and-conquer query minimizer \textsc{minQ} (a modification of \textsc{QuickXPlain} \cite{junker04}) used in e.g.\ \cite{Shchekotykhin2012,Rodler2013} and described comprehensively in \cite[Sec.~8.3]{Rodler2015phd}. However, to additionally meet conditions (2) and (3), we modify the input to \textsc{minQ} accordingly. In particular, we use the following sorting $[Q_{impl},asc_{p}(Q)]$ of formulas in $Q'$ where $[X,Y]$ denotes a list containing first (i.e.\ leftmost) all elements of the collection $X$ and then all elements of the collection $Y$, and $asc_{crit}(X)$ refers to the list of elements in the collection $X$ sorted ascending based on $crit$. 

Then, the application of $\textsc{minQ}$ to the sorted enriched query $[Q_{impl},asc_{p}(Q)]$ returns a query $Q^{*}$ compliant with requirements (1) -- (3) \cite[Cor.~25]{DBLP:journals/corr/Rodler16a}.

\begin{figure*}[t]
	\centering
		\includegraphics[width=\linewidth]{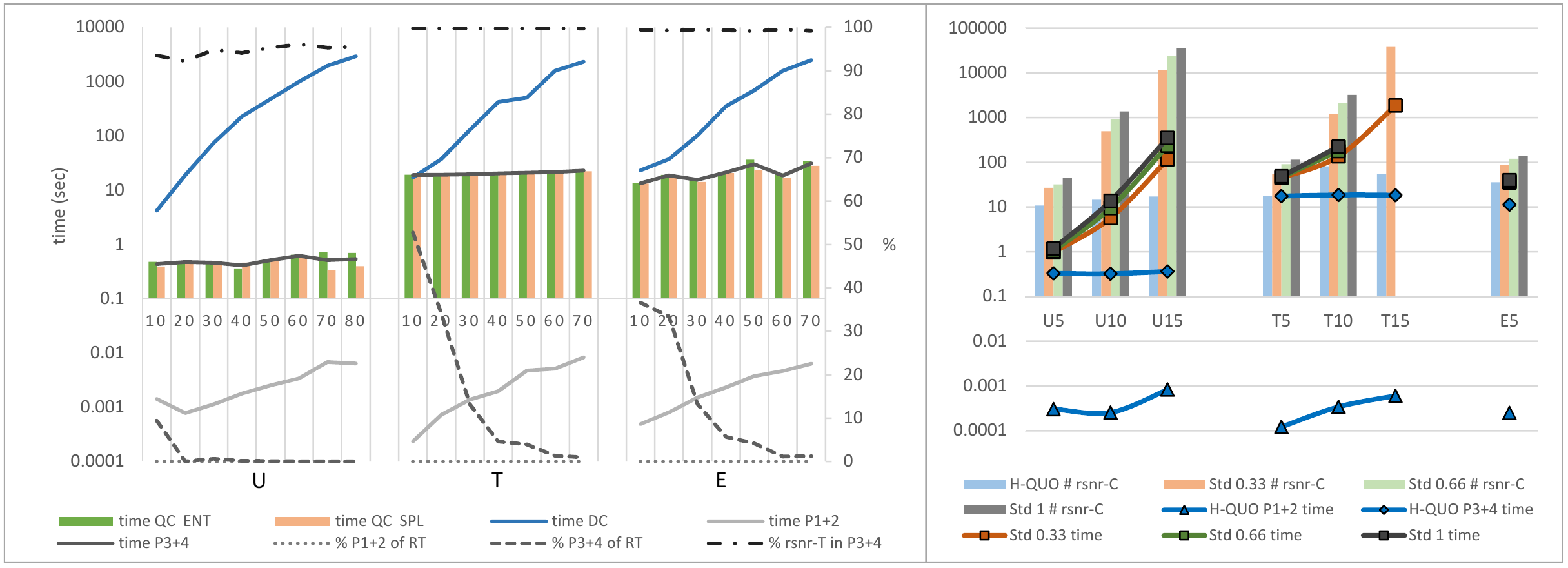}
	\caption{Experiment results for Exp1 (left) and Exp2 (right). All values are 5 trial averages. The x-axes enumerate (KB,$|\mD|$)-configurations. \textbf{(Left):} H-QUO only. Shows times (left \emph{logarithmic} y-axis, continuous lines and bars) for query computation (QC) using $m\in\setof{\mathsf{ENT},\mathsf{SPL}}$, for phases P1+2 and P3+4 (averaged over $\mathsf{ENT},\mathsf{SPL}$) separately and for 
	diagnosis computation (DC), 
	and $\%$ (right y-axis, dashed lines) of debugger reaction time (RT), i.e.\ time between two queries, spent for P1+2 and P3+4 and $\%$ of time used for reasoning (rsnr-T) during P3+4.
	\textbf{(Right):} H-QUO vs.\ standard methods (Std). Times and \# of reasoner calls (rsnr-C) shown by lines and bars, respectively. The 
	scale on the 
	y-axis is logarithmic. In ``Std $i$'' the $i$ denotes the fraction of the QP search space considered. Where no bars or lines are shown, a 1-hour-timeout occurred.
	}
	\label{fig:E1+E2}
\end{figure*}

\section{Evaluation}
We conducted two experiments Exp1 (scalability) and Exp2 (comparison with standard approach) using the faulty real-world KBs University (U), Transportation (T) and Economy (E) from an OWL ontology dataset \cite{Kalyanpur.Just.ISWC07}. We chose these three KBs since their number of minimal diagnoses ($90$, $1782$, $864$) \cite{Shchekotykhin2012}, is large enough for our tests, and since there are sound and complete OWL reasoners like HermiT \cite{Shearer2008} which we used in our tests for consistency checks and the computation of classification and realization entailments \cite{Baader2007}. In both experiments, we used the direct diagnosis generation technique \textsc{Inv-HS-Tree} presented in \cite{Shchekotykhin2014} to generate for each iteration a set of minimal diagnoses $\mD$ where $|\mD| \in \setof{10,20,\dots,90}$ in Exp1 and $|\mD|\in\setof{5,10,15,20}$ in Exp2. To always obtain a different set $\mD$, we incorporated a random re-sorting of the KB-formulas before each call of \textsc{Inv-HS-Tree}, cf.\ \cite{Shchekotykhin2014}. Each test run involved $5$ iterations for fixed KB and $|\mD|$. In each iteration of (a)~Exp1, we used our new approach H-QUO with (heuristics for) one of the query selection strategies split-in-half ($\mathsf{SPL}$) or entropy ($\mathsf{ENT}$) \cite{Shchekotykhin2012}
to generate an optimized query per iteration (i.e.\ performing all four phases P1 -- P4), (b)~Exp2, we used H-QUO just as in Exp1 and additionally applied the standard method (see Related Work), Std for short, to search different random fractions (full, $\frac{2}{3}$, $\frac{1}{3}$) of the QP search space. Note, in each iteration of Exp2, H-QUO and Std were carried out with exactly the same settings.

\subsubsection{Results of Exp1.} 
%
Fig.~\ref{fig:E1+E2} (left side) shows the results for all runs 
where no timeout (execution time $> 1$ hour) occurred in any iteration. For all runs, if a timeout occurred, it was due to diagnosis computation time (see blue lines). On the contrary, the query computation time -- which is almost equal for $\mathsf{SPL},\mathsf{ENT}$ (see bars) -- is give or take independent of $|\mD|$, as indicated by the gray continuous lines. These give the average over $\setof{\mathsf{SPL},\mathsf{ENT}}$ of query computation time needed for P1+2 (\textsc{findQPartition} and \textsc{selectQueryForQPartition}) and P3+4 (\textsc{enrichQuery} and \textsc{optimizeQuery}). We can see that the former is always negligible (never more than $0.03\%$ of overall debugger reaction time, see dotted line) and the latter requires never more than $31.3$ sec (case E70), even though considering QP search space sizes of up to $O(2^{80})$. Of the time consumed by P3+4, reasoning time (polynomial number of calls to a reasoner) never amounted to less than $92\%$, for the most costly cases (KBs T and E) even more than $99\%$ (dashed line at the top). So, the mentioned $31.3$ sec include exactly $31.07$ sec reasoning time. Note, since the execution of P1+2 happens instantaneously (always less than $0.01$ sec) thanks to the reasoner avoidance, the line for P3+4 maps at the same time the overall costs of H-QUO. For increasing $|\mD|$, these have an asymptotically negligible influence on the reaction time of a debugger (decreasing dashed line). Hence, with H-QUO, query computation is not the bottleneck anymore, as opposed to the usage of standard methods (see Exp2). 
In other words, if a set of leading diagnoses $\mD$ of a given cardinality is computable in reasonable time, so is a query (which is already optimized along various dimensions!).

Further, over all runs in Exp1, the average (1)~CQ size was $5.9$, (2)~returned query size was $3.0$ ($\mathsf{ENT}$) and $2.7$ ($\mathsf{SPL}$) formulas, (3)~number of expanded and generated CQPs during P1 was $9.5$ and $85.5$, (4)~number of CQP successors was $10$, (5)~trait size was $1.6$ (very small variance over different (KB,$|\mD|$)-runs), (6)~number of expanded and generated HS nodes during P2 was $3.6$ and $5.1$ (small variance over different (KB,$|\mD|$)-runs), (7)~size reduction of CQ achieved by P2 was $43\%$, (8)~query size increase due to enrichment in P3 was independent of $|\mD|$, but strongly dependent on the KB (number of implicit entailments) and amounted to $14\%$ (U), $2250\%$ (T) and $78200\%$ (factor $783$) (E), (9)~query size reduction through P4 was $13\%$ (U), $96\%$ (T) and $99.8\%$ (E).

\subsubsection{Results of Exp2.} Fig.~\ref{fig:E1+E2} (right side) depicts the results for all runs where no timeout (execution time $> 1$ hour) occurred in any iteration. Unlike in Exp1, here timeouts occurred \emph{only} for Std and due to query computation time (diagnosis computation took never longer than 4 min per iteration). Like in the scalability experiment (Exp1), H-QUO manifests constant behavior despite growing $|\mD|$. We point out that for $|\mD| = 10$, which is already larger than the leading diagnoses size commonly used \cite{Shchekotykhin2012,Rodler2013}, H-QUO is already one order of magnitude (logarithmic, base 10 axis) better than Std 0.33 (see blue diamond and orange lines), even though Std considers only one third of the QP search space in this setting (hence suboptimal queries are possible). Note that the time for P3+4 can again be seen as overall time for H-QUO (see negligible values along blue triangle P1+2 line). For $|\mD| = 15$, the time improvement already exceeds two orders of magnitude, i.e.\ 99\%.
\balance
Whereas no reasoner is required for P1+2, P3+4 exhibits more than one order of magnitude fewer reasoner calls than Std in all runs where $n \geq 10$. The similarity between orange, green and gray lines further reveals that limiting the search space to a constant fraction has almost no influence on the scalability of Std, as the QP search space grows exponentially. Except for the T15 case (where Std 0.33 falls still slightly short of the timeout), if one Std setting fails, all do so. What is more, for the KB E, where reasoning is more costly, none of the Std setups with $|\mD| > 5$ could finish within an hour. In conjunction with the constant execution time of H-QUO throughout the scalability tests (Exp1), this means that H-QUO requires only a fraction of less than $1\%$ of the execution time of Std for all cases where $|\mD| \geq 15$.

%

\section{Conclusions}
In this work we have presented a new query generation and optimization approach. Given a number of diagnoses, it allows for the computation of a query that has (a)~optimal discrimination properties (minimal expected number of subsequent queries to be answered) and (b)~minimal query cardinality or best comprehensibility (minimal effort for the user per query). Importantly, this can be accomplished without any usage of reasoning services. Optionally, however, the latter can be utilized to further enhance an optimized query by replacing potentially complex query elements by simpler surrogates without harming the query's optimal properties. An extensive evaluation using real-world problems testifies the perfect scalability of the approach -- quasi constant time performance for increasing problem size and numbers of diagnoses -- and shows that it drastically outperforms existing methods. In particular, our algorithm outputs an optimal query in less than $0.01$ sec for all tested problems and search space sizes of up to $O(2^{80})$.  

\newpage
\bibliographystyle{aaai} 
\balance
\bibliography{library}

\end{document}